\documentclass[11pt,a4paper]{article}
\usepackage[hyperref]{acl2021}
\usepackage{times}
\usepackage{latexsym}

\usepackage{microtype}
\usepackage{enumitem}
\usepackage{multirow}
\usepackage{indentfirst}
\usepackage{colortbl}
\usepackage{float}
\usepackage{tikz}
\newcommand*{\circled}[1]{\lower.7ex\hbox{\tikz\draw (0pt, 0pt)%
    circle (.5em) node {\makebox[1em][c]{\small #1}};}}

\usepackage{graphicx}
\usepackage{diagbox}

\aclfinalcopy 
\def\aclpaperid{752} 

\title{Improving Formality Style Transfer with Context-Aware Rule Injection}

\author{Zonghai Yao \\
  University of Massachusetts Amherst \\
  \texttt{zonghaiyao@cs.umass.edu} \\\And
  Hong Yu \\
  University of Massachusetts Lowell \\
  \texttt{hong\_yu@uml.edu} \\}

\date{}

\begin{document}
\maketitle
\begin{abstract}
Models pre-trained on large-scale regular text corpora often do not work well for user-generated data where the language styles differ significantly from the mainstream text. Here we present Context-Aware Rule Injection (CARI), an innovative method for formality style transfer (FST). CARI injects multiple rules into an end-to-end BERT-based encoder and decoder model. It learns to select optimal rules based on context. The intrinsic evaluation showed that CARI achieved the new highest performance on the FST benchmark dataset. Our extrinsic evaluation showed that CARI can greatly improve the regular pre-trained models' performance on several tweet sentiment analysis tasks.
\end{abstract}

\section{Introduction}
Many user-generated data deviate from standard language in vocabulary, grammar, and language style. For example, abbreviations, phonetic substitutions, Hashtags, acronyms, internet language, ellipsis, and spelling errors, etc are common in tweets \cite{ghani2019social, muller2019enhancing, han2013lexical, liu2020named}.
Such irregularity leads to a significant challenge in applying existing language models pre-trained on large-scale corpora dominated with regular vocabulary and grammar. One solution is using formality style transfer (FST) \cite{rao-tetreault-2018-dear}, which aims to transfer the input text's style from the informal domain to the formal domain. This may improve the downstream NLP applications such as information extraction, text classification and question answering. 

\begin{figure*}[t]
            \centering
            \includegraphics[width=\linewidth]{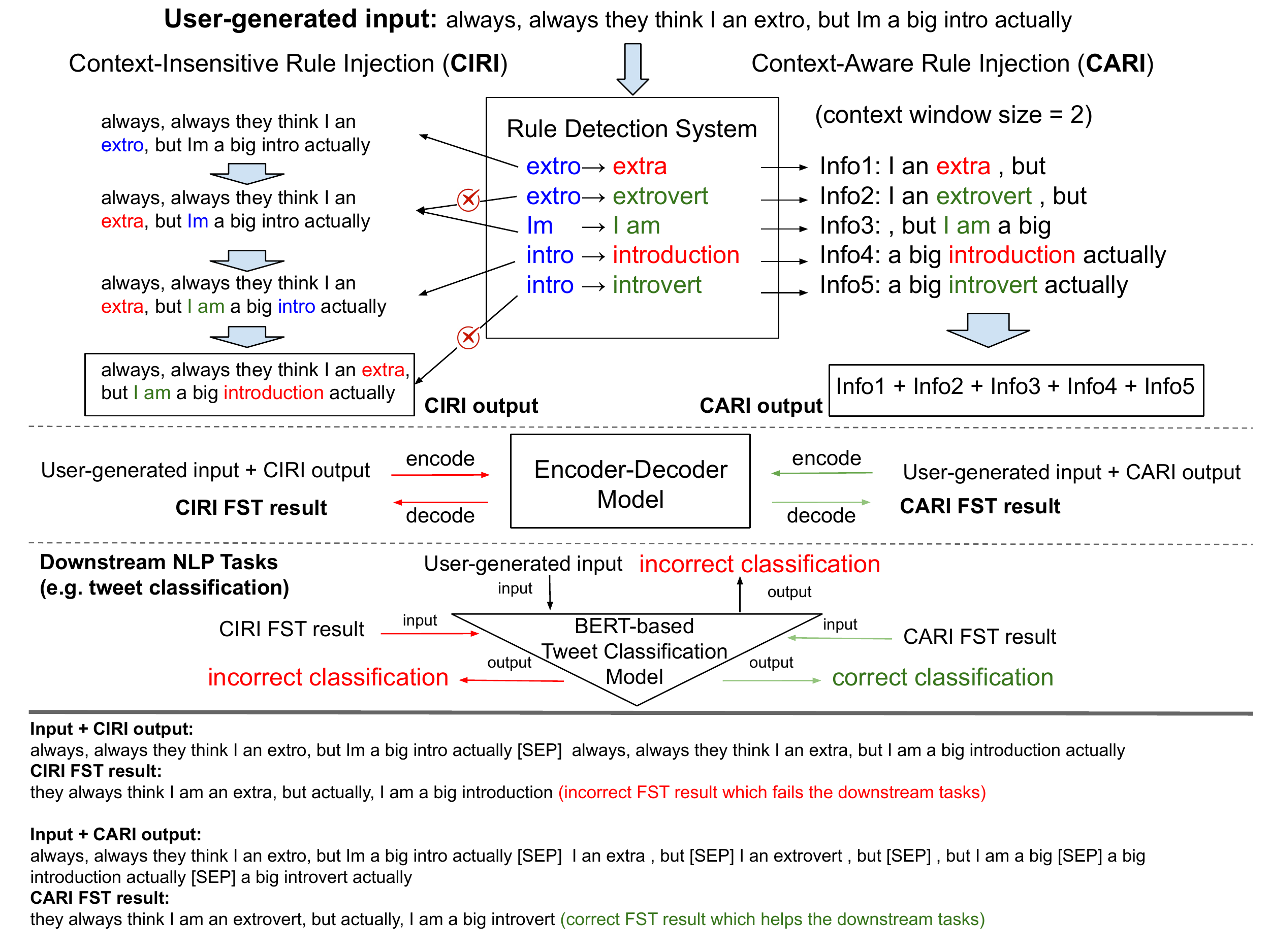}
            \caption{An example of using Context-Insensitive Rule Injection (\textbf{CIRI}) and Context-Aware Rule Injection (\textbf{CARI}) FST models. CIRI models are not context aware and therefore select rules arbitrarily and in this case, apply the rules First Come First Serve (FCFS). The errors introduced ("extra" and "introduction") in the CIRI model impact the downstream NLP tasks, and in this case leading to the incorrect sentiment classification. 
            In CARI, rules are associated with context and through training, CARI can learn to choose the right rules according to the context. This leads to improved FST thereby improves the downstream sentiment classification tasks.}
            \label{description}
    \end{figure*}

A common challenge for FST is low resource \cite{wu2020dataset, malmi2020unsupervised, wang2020formality}. 
Therefore, approaches that integrate external knowledge, such as rules, have been developed. 
However, existing work \cite{rao-tetreault-2018-dear, wang-etal-2019-harnessing} deploy context-insensitive rule injection methods (CIRI). 
As shown in Figure \ref{description}, when we try to use CIRI-based FST as the preprocessing for user-generated data in the sentiment classification task, according to the rule detection system, "extro" has two suggested changes "extra" or "extrovert" and "intro" corresponds to either "introduction" or "introvert." The existing CIRI-based FST models would arbitrarily choose rules following first come first served (FCFS). As such, the input "always, always they think I an extro, but Im a big intro actually" could be translated wrongly as "they always think I am an extra, but actually, I am a big introduction." This leads to the wrong sentiment classification since the FST result completely destroys the original input's semantic meaning.

In this work, we propose Context-Aware Rule Injection (CARI), an end-to-end BERT-based encoder and decoder model that is able to learn to select optimal rules based on context.
As shown in Figure \ref{description}, CARI chooses rules based on context. With CARI-based FST, pre-trained models can perform better on the downstream natural language processing (NLP) tasks. In this case, CARI outputs the correctly translated text "they always think I am an extrovert, but actually, I am a big introvert," which helps the BERT-based classification model have the correct sentiment classification.  

In this study, we performed both intrinsic and extrinsic evaluation of existing FST models and compared them with the CARI model. The intrinsic evaluation results showed that CARI improved the state-of-the-art results from 72.7 and 77.2 to 74.31 and 78.05, respectively, on two domains of a FST benchmark dataset. For the extrinsic evaluation, we introduced several tweet sentiment analysis tasks. Considering that tweet data is typical informal user-generated data, and regular pre-trained models are usually pre-trained on formal English corpora, using FST as a preprocessing step of tweet data is expected to improve the performance of regular pre-trained models on tweet downstream tasks. We regard measuring such improvement as the extrinsic evaluation. The extrinsic evaluation results showed that using CARI model as the prepocessing step improved the performance for both BERT and RoBERTa on several downstream tweet sentiment classification tasks. Our contributions are as follows:

\begin{enumerate}[topsep=0pt,itemsep=0ex,partopsep=0ex,parsep=0ex]
    \item We propose a new method, CARI, to integrate rules for pre-trained language models. CARI is context-aware and can be trained end-to-end with the downstream NLP applications. 
    \item We have achieved new state-of-the-art results for FST on the benchmark GYAFC dataset.
    \item We are the first to evaluate FST methods with extrinsic evaluation 
    and we show that CARI outperformed existing rule-based FST approaches for sentiment classification.
\end{enumerate}

\section{Related work}

\paragraph{Rule-based Formality Style Transfer}
In the past few years, style-transfer generation has attracted increasing attention in NLP research. Early work transfers between modern English and
the Shakespeare style with a phrase-based machine translation system  \cite{xu2012paraphrasing}. Recently, style transfer has been more recognized as a controllable text generation problem \cite{hu2017toward}, where the style may be designated as sentiment \cite{fu2018style}, tense \cite{hu2017toward}, or even general syntax \cite{bao2019generating, chen2019controllable}. Formality style transfer has been mostly driven by the Grammarly’s Yahoo Answers Formality Corpus (GYAFC) \cite{rao-tetreault-2018-dear}. Since it is a parallel corpus, FST usually takes a seq2seq-like approach \cite{niu2018multi, xu2019formality}. Existing research attempts to integrate the rules into the model because the GYAFC is low resource. However, rule matching and selection are context insensitive in previous methods \cite{wang-etal-2019-harnessing}. This paper focuses on developing methods for context-aware rule selection.

\paragraph{Evaluating Style Transfer}
Previous work on style transfer \cite{xu2012paraphrasing, jhamtani2017shakespearizing, niu2017study, sennrich2016controlling} has re-purposed the machine translation metric BLEU \cite{papineni2002bleu} and the paraphrase metric PINC \cite{chen2011collecting} for evaluation. \citet{xu2012paraphrasing} introduced three evaluation metrics based on cosine similarity, language model and logistic regression. They also introduced human judgments for adequacy, fluency and style \cite{xu2012paraphrasing, niu2017study}.  \citet{rao-tetreault-2018-dear} evaluated formality, fluency and meaning on the GYAFC dataset. Recent work on the GYAFC dataset \cite{wang-etal-2019-harnessing, zhang2020parallel} mostly used BLEU as the evaluation metrics for FST. 
However, all aforementioned work focused on intrinsic evaluations. Our work has in addition evaluated FST extrinsically for downstream NLP applications. 

\paragraph{Lexical Normalisation}
Lexical normalisation \cite{han2011lexical, baldwin2015shared} is the task of translating non-canonical words into canonical ones. Like FST, lexical normalisation can also be used to preprocess user-generated data. 
The MoNoise model \cite{van2017monoise} is a state-of-the-art model based on feature-based Random Forest. 
The model ranks candidates provided by modules such as a spelling checker (aspell), a n-gram based language model and word embeddings trained on millions of tweets. Unlike FST, MoNoise and other lexical normalisation models can not change data’s language style. In this study, we explore the importance of language style transfer for user-generated data by comparing the results of MoNoise and FST models on tweets NLP downstream tasks.

\paragraph{Improving language models' performance for user-generated data}
User-generated data often deviate from standard language. In addition to the formality style transfer, there are some other ways to solve this problem \cite{eisenstein2013bad}. Fine-tuning on downstream tasks with a user-generated dataset is most straightforward, but this is not easy for many supervised tasks without a large amount of accurately labeled data. Another method is to fine-tune pre-trained models on the target domain corpora \cite{gururangan2020don}. However, it also requires sizable training data, which could be resource expensive \cite{sohoni2019low, dai2019transformer, yao2020zero}.



\section{Approach}

For the downstream NLP tasks where input is user-generated data, we first used the FST model for preprocessing, and then fine-tuned the pre-trained models (BERT and RoBERTa) with both the original data $D_{ori}$ and the FST data $D_{FST}$, which were concatenated with a special token $[SEP]$, forming an input like $(D_{ori} [SEP] D_{FST})$. 

For the formality style transfer task, we use the BERT-initialized encoder paired with the BERT-initialized decoder \cite{rothe2020leveraging} as the Seq2Seq model. All weights were initialized from a public BERT-Base checkpoint \cite{devlin-etal-2019-bert}. The only variable that was initialized randomly is the encoder-decoder attention. Here, we describe CARI and several baseline methods of injecting rules into the Seq2Seq model.

\subsection{No Rule (NR)}
First we fine-tuned the BERT model with only the original user-generated input. Given an informal input $x_i$ and formal output $y_i$, we fine-tuned the model with $\{(x_i, y_i)\}^{M}_{i=0}$, where M is the number of data.

\subsection{Context Insensitive Methods}
For baseline models, we experimented with two state-of-the-art methods for injecting rules. We followed \citet{rao-tetreault-2018-dear} to create a set of rules to convert original data $x_i$ to prepossessed data $x^{\prime}_i$ by rules, and then fine-tune the model with parallel data $\{(x^{\prime}_i, y_i)\}^{M}_{i=0}$. This is called \textbf{Rule Base (RB)} method. The prepossessed data, however, serves as a Markov blanket, i.e., the system is unaware of the original data, provided that only the prepossessed one is given. Therefore, the rule detection system could easily make mistakes and introduce noise.

\citet{wang-etal-2019-harnessing} improved the RB by concatenating the original text $x_i$ with the text processed by rules $x^{\prime}_i$ with a special token $[SEP]$ in between, forming a input like $(x_i \; [SEP] \; x^{\prime}_{i})$. In this way, the model can make use of a rule detection system but also recognize its errors during the fine-tuning. This is called \textbf{Rule Concatenation (RCAT)} method. However, both RB and RCAT methods are context insensitive, the rules were selected arbitrarily. In Figure \ref{description} CIRI part, "extra" and "introduction" were incorrectly selected. This greatly limits the performance of the rule-based methods.

\subsection{Context-Aware Rule Injection (CARI)}

As shown in Figure \ref{description}, the input of CARI consists of the original sentence $x_i$ and supplementary information. Suppose that $r_i$ is an exhaustive list of the rules that are successfully matched on $x_i$. 
We make $r_i = \{(t_{i, j}, c_{i, j}, a_{i, j})\}^{N}_{j=0}$, where N is the total number of matched rules in $r_i$. Here, $t_{i, j}$ and $c_{i, j}$ are the corresponding matched text and context in the original sentence, respectively, for every matched rule in $r_i$, and $a_{i, j}$ are the corresponding alternative texts for every matched rule in $r_i$. 
Each supplementary information is composed of one alternative text $a_{i, j}$ and its corresponding context $c_{i, j}$. We connect all the supplementary information with the special token $[SEP]$ and then connect it after the original input. In this way, we form an input like $(x_i \; [SEP] \; a_{i,1}, c_{i,1} \; [SEP] ... \; [SEP] \; a_{i,j}, c_{i,j})$. Finally, the concatenated sequence and the corresponding formal reference $y_i$ serve as a parallel text pair to fine-tune the Seq2Seq model. Like RCAT, CARI can also use rule detection system and recognize its errors during the fine-tuning. Furthermore, since we keep all rules in the input, CARI is able to dynamically identify which rule to use, maximizing the use of the rule detection system.

\section{Experimental setup}

\subsection{Datasets}

For the intrinsic evaluation, we used the GYAFC dataset.\footnote{https://github.com/raosudha89/GYAFC-corpus} 
It consists of handcrafted informal-formal sentence pairs in two domains, namely, Entertainment \& Music (E\&M) and Family \& Relationship (F\&R). 
Table \ref{dataset} shows the statistics of the training, validation, and test sets for the GYAFC dataset. In the validation and test sets of GYAFC, each sentence has four references. For better exploring the data requirements of different methods to combine rules, we followed \citet{zhang2020parallel} and used the back translation method \cite{sennrich-etal-2016-neural} to obtain additional 100,000 data for training. For rule detection system, we used the grammarbot API,\footnote{https://www.grammarbot.io/}, and Grammarly\footnote{https://www.grammarly.com/} to help us create a set of rules. 

For the extrinsic evaluation, we used two datasets for sentiment classification: SemEval-2018 Task 1: Affect in Tweets EI-oc \cite{mohammad2018semeval}, and Task 3: Irony Detection in English Tweets \cite{van2018semeval}. Table \ref{dataset} shows the statistics of the training, validation, and test set for the two datasets.
We normalized two tweet NLP classification datasets by translating word tokens of user mentions and web/url links into special tokens @USER and HTTPURL, respectively, and converting emotion icon tokens into corresponding strings. 

    \begin{table}
        \centering
        \begin{tabular}{c|c|c|c}
        \hline
        \small{FST GYAFC dataset} & Train & Valid & Test  \\
        \hline
        \small{Entertainment \& Music} & 52,595 & 2,877 & 1,416  \\
        \hline
        \small{Family \& Relationship} & 51,967 & 2,788 & 1,322  \\
        \hline
        \multicolumn{4}{l}{}\\
        \hline
        \small{Affect in Tweets EI-oc} & Train & Valid & Test  \\
        \hline
        \small{anger} & 1,701 & 388 & 1,002  \\
        \hline
        \small{fear} & 2,252 & 389 & 986  \\
        \hline
        \small{joy} & 1,616 & 290 & 1,105  \\
        \hline
        \small{sadness} & 1,533 & 397 & 975  \\
        \hline
        \multicolumn{4}{l}{}\\
        \hline
        \small{Irony Detection} & Train & Valid & Test  \\
        \hline
        \small{Irony-a} & 3067 & 767 & 784  \\
        \hline
        \small{Irony-b} & 3067 & 767 & 784  \\
        \hline
        \end{tabular}
        \caption{The data statistics for GYAFC dataset of Formality style transfer task and Tweet NLP downstream classification datasets.}
        \label{dataset}
    \end{table}
    
    \begin{figure*}[t]
            \centering
            \includegraphics[width=\linewidth]{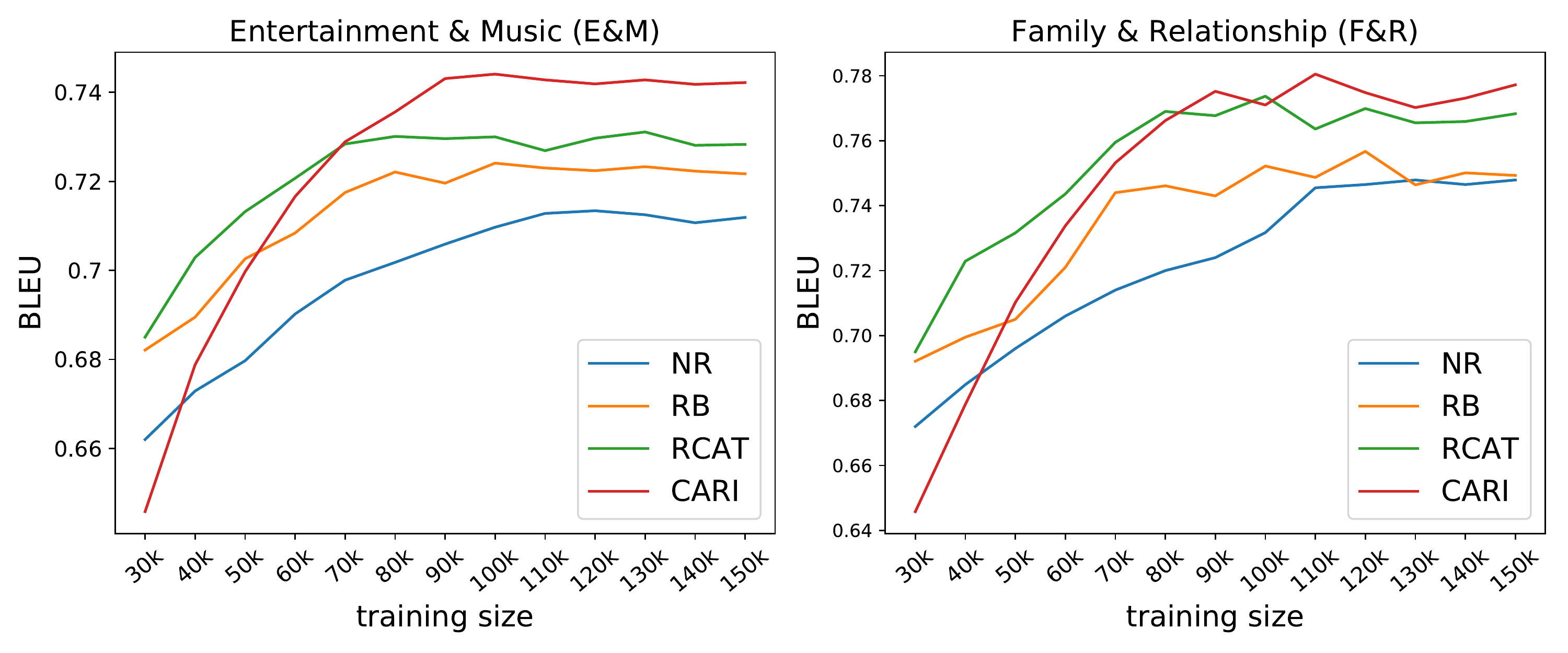}
            \caption{The performance (BLEU) of different rule injection methods with different training size. The results show that:
            1) CARI achieved the best results in both E\&M and F\&R domains. 
            2) Rb, RCAT and CARI achieved optimal performance on less training size compared with NR, indicating the advantages of integrating rules to mitigate the low resource challenge. 
            3) compared with Rb and RCAT, CARI required slightly larger training size due to its context-aware learning model.}
            \label{main}
    \end{figure*}

\subsection{Fine-tuning models}
We employed the transformers library \cite{wolf2019huggingface} to independently fine-tune the BERT-based encoder and decoder model for each method in 20,000 steps (intrinsic evaluation), and fine-tune the BERT-based and RoBERTa-based classification models for each tweet sentiment analysis task in 10,000 steps (extrinsic evaluation). We used the Adam algorithm \cite{kingma2014adam} to train our model with a batch size 32. We set the learning rate to 1e-5 and stop training if validation loss increases in two successive epoch. We computed the task performance every 1,000 steps on the validation set. Finally, we selected the best model checkpoint to compute the performance score on the test set. We repeated this fine-tuning process three times with different random seeds and reported each final test result as an average over the test scores from the three runs. During inference, we use beam search with a beam size of 4 and beam width of 6 to generate sentences. The whole experiment is carried out on 1 TITANX GPU. Each FST model finished training within 12 hours.

\subsection{Intrinsic Evaluation Baselines}
We used two state-of-the-art models, which were also relevant to our methods, as the strong intrinsic baseline models.

\paragraph{ruleGPT}
Like RCAT, \citet{wang-etal-2019-harnessing} aimed to solve the problem of information loss and noise caused by directly using rules as normalization in preprocessing. They put forward the GPT \cite{radford2019language} based methods to concatenate the original input sentence and the sentence preprocessed by the rule detection system. Like the CIRI methods (RB, RCAT), their methods could not make full use of rules since they were also context-insensitive when selecting rules. 

\paragraph{BT + M-Task + F-Dis}
\citet{zhang2020parallel} used three data augmentation methods, Back translation \cite{sennrich-etal-2016-neural}, Formality discrimination, and Multi-task transfer to solve the low-resource problem. In our experiments, we also use the back translation method to obtain additional data because we want to verify the impact on the amount of training data required when using different methods to combine rules.

\subsection{Extrinsic Evaluation Baselines}

    \begin{table*}
        \centering
        \begin{tabular}{l|c|cccc|cccc}
        \hline

        \multicolumn{10}{c}{Irony Detection (evaluation metrics: F1)} \\
        \hline
        & \small{UCDCC} & \small{BERT} & \small{MoNoise} & \small{RCAT} & \small{CARI} & \small{RoBERTa} & \small{MoNoise} & \small{RCAT} & \small{CARI} \\  
        \hline
        Irony-a & 72.4 & 71.8 & 72.2 & 72.2 & 72.5 & 72.6 & 72.6 & 73.1 & \textbf{73.7} \\ 
        
        Irony-b & 50.7 & 48.6 & 48.8 & 50.2 & 50.9 & 51.2 & 51 & 53.3 & \textbf{53.8} \\ 
        \hline
        \multicolumn{10}{c}{}\\
        \hline
        \multicolumn{10}{c}{Affect in Tweets EI-oc (evaluation metrics: Pearson r)} \\
        \hline
        & \small{SeerNet} & \small{BERT} & \small{MoNoise} & \small{RCAT} & \small{CARI} & \small{RoBERTa} & \small{MoNoise} & \small{RCAT} & \small{CARI}  \\ 
        \hline
        
        Joy & 72 & 69.1 & 68.6 & 69.7 & 70.4 & 71.8 & 71.5 & 72.9 & \textbf{73.5} \\ 
        
        Anger & 70.6 & 71.6 & 71.7 & 71.9 & 72 & 72 & 71.7 & 72.3 & \textbf{72.2} \\ 
        
        Sad & \textbf{71.7} & 66.8 & 66.4 & 67.4 & 68.3 & 68.2 & 68 & 69.1 & 70.1 \\ 
        
        Fear & 63.7 & 66.9 & 66.8 & 67.1 & 69.2 & 69.8 & 69.4 & 70.5 & \textbf{71.4} \\ 
        
        \hline
        \end{tabular}
        \caption{The extrinsic evaluation results on tweet sentiment analysis tasks. Through observation, we can find that 
        1) Compared with the previous state-of-the-art results, the results of using BERT and RoBERTa directly were often very poor.
        2) Monoise can not effectively improve the results of BERT and RoBERTa, while FST method can.
        3) Compared with RCAT, CARI can better improve the results of BERT and RoBERTa on user-generated data.}
        \label{tweetsTask}
    \end{table*}

BERT \cite{devlin2018bert} and RoBERTa \cite{liu2019roberta} are two typical regular language models pre-trained on large-scale regular formal text corpora, like BooksCorpus \cite{zhu2015aligning} and English Wikipedia. The user-generated data, such as tweets, deviate from the formal text in vocabulary, grammar, and language style. As a result, regular language models often perform poorly on user-generated data. FST aims to generate a formal sentence given an informal one, while keeping its semantic meaning. A good FST result is expected to make regular language models perform better on user-generated data. 
For the extrinsic evaluation, we chose BERT and RoBERTa as the basic model. We introduced several tweet sentiment analysis tasks to explore the FST models' ability to transfer the user-generated data from the informal domain to the formal domain. Ideally, FST results for tweet data can improve the performance of BERT and RoBERTa on tweet sentiment analysis tasks. We regard measuring such improvement as the extrinsic evaluations.
Besides, tweet data have much unique information, like Emoji, Hashtags, ellipsis, etc., which are not available in the GYAFC dataset. So in the extrinsic evaluation result analysis, although the final scores of FST-BERT and FST-RoBERTa were good, we paid more attention to the improvement of their performance before and after using FST, rather than the scores.

We used two different kinds of state-of-the-art methods as our extrinsic evaluation baselines.

\paragraph{SeerNet and UCDCC}
We used the best results in the SemEval-2018 workshop as the first comparison method. For the task Affect in Tweets EI-o, the baseline is SeerNet \cite{duppada2018seernet}, and for the task Irony Detection in English Tweets, the baseline is UCDCC \cite{ghosh2018ironymagnet}.

\paragraph{MoNoise}
MoNoise \cite{van2017monoise} is the state-of-the-art model for the lexical normalization \cite{baldwin2015shared}, which aimed to translate non-canonical words into canonical ones. Like the FST model, MoNoise can also be used as the prepossessing step in tweet classification tasks to normalize tweet input. So we used MoNoise as another comparison method.

\section{Experimental results}

\subsection{Intrinsic Evaluation}

    \begin{table}
        \centering
        \begin{tabular}{l|cc}
        \hline
        
        \multirow{2}*{Model} & E\&M & F\&R \\
        
        & BLEU & BLEU \\
        
        \hline
        no edit & \small{50.28} & \small{51.67}\\
        
        \hline
        
        ruleGPT & \small{72.7} & \small{77.26} \\
        
        BT + M-Task + F-Dis & \small{72.63} & \small{77.01} \\
        
        \hline
        NR & \small{71.94} & \small{75.65} \\
        
        RB & \small{72.01} & \small{75.67} \\
        
        RCAT & \small{73.01} & \small{77.37} \\
        
        CARI & \textbf{\small{74.31}} & \textbf{\small{78.05}} \\
        
        \hline
        \end{tabular}
        \caption{The comparison of our approaches to the state-of-the-art results on the GYAFC test set.}
        \label{baselines}
    \end{table}

Figure \ref{main} showed the validation performance on both the E\&M and the F\&R domain. Compared to the NR, the RB did not significantly improve. As we discussed above, even though the rule detection system will bring some useful information, it will also make mistakes and introduce noise. RB has no access to the original data, so it cannot distinguish helpful information from noise and mistakes. On the contrary, both RCAT and CARI have access to the original data, so their results improved a lot compared with RB. 
CARI had a better result compared to the RCAT. This is because RCAT is context insensitive while CARI is context-aware when selecting rules to modify the original input. Therefore, CARI is able to learn to select optimal rules based on context, while RCAT may miss using many correct rules with its pipeline prepossessing step for rules.

Figure \ref{main} also showed the relationship between the different methods and the different training size. Compared with the NR method, the three methods which use rules can reach their best performance with smaller training size. This result showed the positive effect of adding rules in the low-resource situation of the GYAFC dataset. Moreover, CARI used larger training set to reach its best performance than RB and RCAT, since it needed more data to learn how to dynamically identify which rule to use.

    \begin{table}
        \centering
        \begin{tabular}{l|cccccc}
        \hline
        
        & \multicolumn{6}{c}{context window size for CARI}  \\
        
        & 0 & 1 & 2 & 3 & 4 & 5\\
        
        \hline
        \small{E\&M} & \small{68.1} & \small{72.5} & \small{74.2} & \small{74.6} & \small{74.3} & \small{74.5}\\
        
        \small{F\&R} & \small{70.5} & \small{74.3} & \small{76.9} & \small{77.5} & \small{76.8} & \small{77.3}\\

        \hline
        \end{tabular}
        \caption{CARI performance (BLEU) by different context window size. When the context window size reach 2, the model can make good use of the rules' information.}
        \label{window_size}
    \end{table}
    
In Table \ref{window_size}, we explored how large the context window size was appropriate for the CARI method on GYAFC dataset. The results showed that for both domains when the window size reaches two (taking two tokens each from the text before and after), Seq2Seq model can well match all rules with the corresponding position in the original input and select the correct one to use.

    \begin{table*}
        \centering
        \begin{tabular}{lll}

        \hline
        Example 1: &
        Source: & explain 2 ur parents that u really want 2 act !!! \\
        & MoNoise: & explain to your parents that you really want to act !\\
        & FST: & explain to your parents that you want to act . \\
        
        \hline
        
        Example 2: &
        Source: & my observation skills??? wow, very dumb...... \\
        & MoNoise: & my observation skills ? wow, very dumb . very\\
        & FST: & my observation skills are very bad . \\
        
        \hline
        
        Example 3: &
        Source: & hell no your idiodic for asking . \\
        & NR: & i do not understand your question . \\
        & CARI: & absolutely not and i feel you are idiotic for asking .\\
        
        \hline

        Example 4: &
        Source: & got exo to share, concert in hk ! u interested ?\\
        & RCAT: & have you got exo to share, concert in hong kong . are you interested ?\\
        & CARI: & i got extra to share , concert in hong kong . are you interested ?\\

        \hline
        Example 5: &
        Source: & fidy cent he is fine and musclar \\
        &Target: & 50 Cent is fine and muscular .\\
        &CARI: & fidy cent is fine and muscular .\\
        
        \hline
        
        Example 6: &
        Source: & if my pet bird gets too flappy, my pet kitty cat might eaty \\
        & Target: & if my pet bird gets too flappy, my pet kitty cat might eat it\\
        & CARI: & if my pet bird gets too flappy, my pet kitty cat might eat me\\
        \hline
        
        \end{tabular}
        \caption{Sample model outputs. Example 1 shows that both MoNoise and FST models can handle some simplest modifications. Example 2 shows that FST can transform the language style of user-generated data, while MoNoise can not. Example 3 shows that NR-based FST can not understand the source because of OOV noises in the data, while CARI-based FST can understand with rules. Example 4 shows the importance of context for rule selection. The word "concern" provides the required context to understand that "exo" refers to an "extra" ticket. In example 5, the rule detection system did not provide the information that the "fidy center" should be "50 Cent (American rapper)", so CARI makes the wrong result. In example 6, CARI mistakenly selected the rule "eat me."}
        \label{example}
    \end{table*}

\subsection{Extrinsic Evaluation}
    
Table \ref{tweetsTask} showed the effectiveness of using the CARI as the preprocessing step for user-generated data on applying regular pre-trained models (BERT and RoBERTa) on the downstream NLP tasks. 

Compared with the previous state-of-the-art results (UCDCC and SeerNet), the results of using BERT and RoBERTa directly were often very poor, since BERT and RoBERTa were only pre-trained on regular text corpora. Tweet data has the very different vocabulary, grammar, and language style from the regular text corpora, so it is hard for BERT and RoBERTa to have good performance with small amount of fine-tuning data.

The results of RCAT and CARI showed that FST can help BERT and RoBERTa improve their performance on tweet data, because they can transfer tweets into more formal text while keeping the original intention as much as possible. CARI performed better than RCAT, which was also in line with the results of intrinsic evaluation. This result also showed the rationality of our extrinsic evaluation metrics.

Comparing the results of MoNoise with BERT and RoBERTa, the input prepossessed by MoNoise can not help the pre-trained model to improve effectively. 
We think that this is because the lexical normalization models represented by MoNoise only translate non-canonical words on tweet data into canonical ones. Therefore, MoNoise can basically solve the problem of different vocabulary between regular text corpora and user-generated data, but it can not effectively solve the problem of different grammar and language style. As a result, for BERT and RoBERTa, even though there is no Out-of-Vocabulary (OOV) problem in the input data processed by MoNoise, they still can not accurately understand the meaning of the input.

This result confirmed the previous view that lexical normalization on tweets is a lossy translation task \cite{owoputi2013improved, nguyen2020bertweet}. On the contrary, the positive results of the FST methods also showed that FST is more suitable as the downstream task prepossessing step of user-generated data. Because FST models need to transfer the informal language style to a formal one while keeping its semantic meaning, which makes a good FST model can ideally handle all the problems from vocabulary, grammar, and language style. This can help most language models pre-trained on the regular corpus, like BERT and RoBERTa, perform better on user-generated data.

\subsection{Manual Analysis}
The prior evaluation results reveal the relative performance differences between approaches. Here, we identify trends per and between approaches. We sample 50 informal sentences total from the datasets and then analyze the outputs from each model. We present several representative results in Table \ref{example}.

Examples 1 and 2 showed that, for BERT and RoBERTa, FST models are more suitable for preprocessing user-generated data than lexical normalization models. In example 1, both methods can effectively deal with the problem at the vocabulary level ("2" to "to," "ur" to "your," and "U" to "you"). However, in example 2, FST can further transform source data into a more familiar language style for BERT and RoBERTa, which is not available in the current lexical normalization methods such as MoNoise.

Example 3 showed the importance of injecting rules into the FST models. The word "idiodic" is a misspelling of "idiotic," which is an OOV. Therefore, without the help of rules, the model can not understand the source data's meanings and produced the wrong final output "I do not understand your question." 

Example 4 showed the importance of context for rule selection. The word "concern" provides the required context to understand that "exo" refers to an "extra" ticket. So the CARI-based model can choose the right one ("exo" to "extra").

Examples 5 and 6 showed the shortcomings of CARI. In example 5, the rule detection system did not provide the information that the "fidy center" should be "50 Cent (American rapper)", so CARI delivered the wrong result. Even though CARI helps mitigate the data low resource challenge, it faces the challenge on its own. CARI depends on the quality of the rules, and in this case, no rule exists that links "fidy" to "50."
In example 6, CARI mistakenly selected the rule "eat me," but not "eat it." This example also demonstrates the data sparsity that CARI faces. Here "eat me" is more commonly used than "eat it."

\section{Conclusions}
In this work, we proposed the Context-Aware Rule Injection(CARI), an innovative method for formality style transfer (FST) by injecting multiple rules into an end-to-end BERT-based encoder and decoder model.
The intrinsic evaluation showed our CARI method achieved the highest performance with previous metrics on the FST benchmark dataset. 
Besides, we were the first to evaluate FST methods with extrinsic evaluation and specifically on the sentiment classification tasks. The extrinsic evaluation results showed that using the CARI-based FST as the preprocessing step outperformed existing rule-based FST approaches. Our results showed the rationality of adding such extensive evaluation.

\bibliographystyle{acl_natbib}
\bibliography{anthology,acl2021,my}


\end{document}